\documentclass[conference]{IEEEtran}
\IEEEoverridecommandlockouts
\usepackage{cite}
\usepackage{amsmath,amssymb,amsfonts}
\usepackage{algorithmic}
\usepackage{graphicx}
\usepackage{textcomp}
\usepackage{xcolor}
\usepackage{float} 
\usepackage{multirow}
\usepackage{hyperref}
\usepackage{hhline}
\usepackage{times}

\def\BibTeX{{\rm B\kern-.05em{\sc i\kern-.025em b}\kern-.08em
    T\kern-.1667em\lower.7ex\hbox{E}\kern-.125emX}}
\begin{document}

\title{Pruning-Based TinyML Optimization of Machine Learning Models for Anomaly Detection in Electric Vehicle Charging Infrastructure}

    









\author{
    \IEEEauthorblockN{Fatemeh Dehrouyeh, Ibrahim Shaer, Soodeh Nikan, Firouz Badrkhani Ajaei, and Abdallah Shami}
    \IEEEauthorblockA{
        \textit{Department of Electrical and Computer Engineering} \\
        \textit{University of Western Ontario} \\
        London, Ontario, Canada
    }
    \IEEEauthorblockA{
        \{fdehrouy, ishaer, snikan, fajaei, Abdallah.Shami\}@uwo.ca
    }
}

\maketitle

\begin{abstract}
With the growing need for real-time processing on IoT devices, optimizing machine learning (ML) models' size, latency, and computational efficiency is essential. This paper investigates a pruning method for anomaly detection in resource-constrained environments, specifically targeting Electric Vehicle Charging Infrastructure (EVCI). Using the CICEVSE2024 dataset, we trained and optimized three models---Multi-Layer Perceptron (MLP), Long Short-Term Memory (LSTM), and XGBoost---through hyperparameter tuning with Optuna, further refining them using SHapley Additive exPlanations (SHAP)-based feature selection (FS) and unstructured pruning techniques. The optimized models achieved significant reductions in model size and inference times, with only a marginal impact on their performance. Notably, our findings indicate that, in the context of EVCI, pruning and FS can enhance computational efficiency while retaining critical anomaly detection capabilities. 
\end{abstract}


\section{Introduction}

Efficiency in deep learning (DL) is crucial due to several challenges. Initially, training and deploying models on a large scale can be costly, consuming significant server resources and contributing to the substantial carbon footprint of data centers. Moreover, for applications requiring real-time processing on IoT and smart devices, models must be optimized to operate effectively within these constrained environments, addressing concerns of privacy, connectivity, and responsiveness. Also, the sensitivity of user data requires that models be capable of learning from minimal data to reduce the need for extensive data collection. Consequently, as the demand for personalized and diverse applications grows, the ability to train and deploy multiple models efficiently without excessively burdening the infrastructure becomes essential, demonstrating the importance of developing more refined, resource-efficient DL strategies~\cite{menghani2023efficient}.

As examples of efficiency in practice, healthcare systems use ML efficiency through federated learning algorithms that monitor chronic diseases in real-time using medical IoT devices, such as wearable sensors and mobile health applications~\cite{patil2023federated}. In computer vision applications, techniques like cluster pruning address the high memory and computational demands of Convolutional Neural Networks (CNNs), enabling real-time algorithms on low-cost IoT devices~\cite{gamanayake2020cluster}. Similarly, in autonomous systems, activation pruning selectively reduces activation maps in early neural network layers, minimizing data movement and computational load while preserving real-time processing capabilities~\cite{samal2020attention}.

Recent advancements in ML have focused on developing techniques that reduce model complexity and resource consumption without compromising performance. Lightweight models, which are designed to maintain high accuracy with fewer computational demands, have emerged as a key solution~\cite{10.1145/3657282}. Pruning is an effective strategy for making lightweight models, which reduces the number of weights in a neural network without significantly affecting performance~\cite{lecun1989optimal}. 

This study emphasizes model optimization through pruning to achieve efficient performance without compromising accuracy. By creating lightweight, resource-efficient models, pruning facilitates deployment on low-power, resource-constrained devices, enabling these devices to support TinyML applications. TinyML enables real-time, on-device anomaly detection, eliminating the need for continuous cloud connectivity~\cite{abadade2023comprehensive}. In the context of EVCI, TinyML can be deployed on a microcontroller within EVs to monitor the Controller Area Network (CAN) bus, providing an effective intrusion detection system~\cite{10620263}. For these applications, the model must be highly efficient to fit within the limited memory and processing capacities of IoT devices, underscoring the importance of pruning techniques to enhance model efficiency.

By the term efficiency, the footprint metrics of a ML model—such as model size, latency, and the number of epochs to convergence—are considered alongside the quality of the model, measured by metrics such as accuracy, precision, and recall~\cite{menghani2023efficient}. As neural networks increase in size with additional layers and nodes, minimizing their storage requirements and computational costs becomes essential, especially for real-time applications~\cite{cheng2018model}. Compressing neural networks not only enhances memory and energy efficiency but also reduces network bandwidth and processing time, and improves privacy. Studies have shown that large networks are often over-parameterized and can be compressed by up to 100 times without significantly impacting accuracy~\cite{ullrich2017soft}.

To the best of our knowledge, there is no previous work on the application of pruning techniques for EVCI. 
Optimizing ML models for EVCI is particularly valuable as it enables real-time intrusion detection systems (IDS) to be installed directly on electric vehicles (EVs). Since EVs are inherently power-constrained devices, it is essential to use optimized ML models that can operate efficiently without compromising battery life. Deploying IDS on EVs enhances their ability to detect and respond to security threats in real time, ensuring the safety and reliability of both the vehicles and the broader charging infrastructure.
The main contributions of this study are outlined as follows:
\begin{itemize}
    \item Using the CICEVSE2024 dataset, we assessed model efficiency and established a benchmark for future studies in EVCI.
    \item By exploring pruning and FS as methods to optimize ML models for anomaly detection in EVCI, we achieved significant reductions in model size and inference time without compromising performance.
    \item Through substantial computational gains, this work paves the way for real-time, on-device anomaly detection in EVCI, supporting the broader adoption of TinyML in this field.
\end{itemize}

The structure of this paper is as follows: Section II covers Data Overview and Preparation Steps, including Dataset Description and Preprocessing; Section III presents the Methodology, detailing Model Training and Hyperparameter Optimization (HPO), FS, and Model Pruning; Section IV discusses the Results and Analysis; and finally, Section V concludes with the Conclusion.

\section{Data Overview and Preparation Steps}

\subsection{Dataset Description}
The EVCI includes several essential components, such as the EV, the Electric Vehicle Charging Station (EVCS), and the Central Management System (CMS). An EVCS, which may consist of one or more Electric Vehicle Supply Equipment (EVSE) units, provides the necessary power to charge EVs. The CMS functions as a cloud-based server that monitors and manages the operations of various charging stations~\cite{10620263}.

Recognized for its utility in cybersecurity research, CICEVSE2024 provides a rich and diverse feature set for training lightweight ML models, making it particularly suitable for this work.
The CICEVSE2024 dataset includes power consumption data, network traffic, and host activities of the EVSE under both benign and attack conditions~\cite{unbdataset, buedi2024enhancing}. 
Despite the availability of power consumption and host activities data, this paper focuses solely on using network traffic data to detect anomalies and classify different attack scenarios.

\begin{figure}[ht]
\centering
\includegraphics[width=0.5\textwidth]{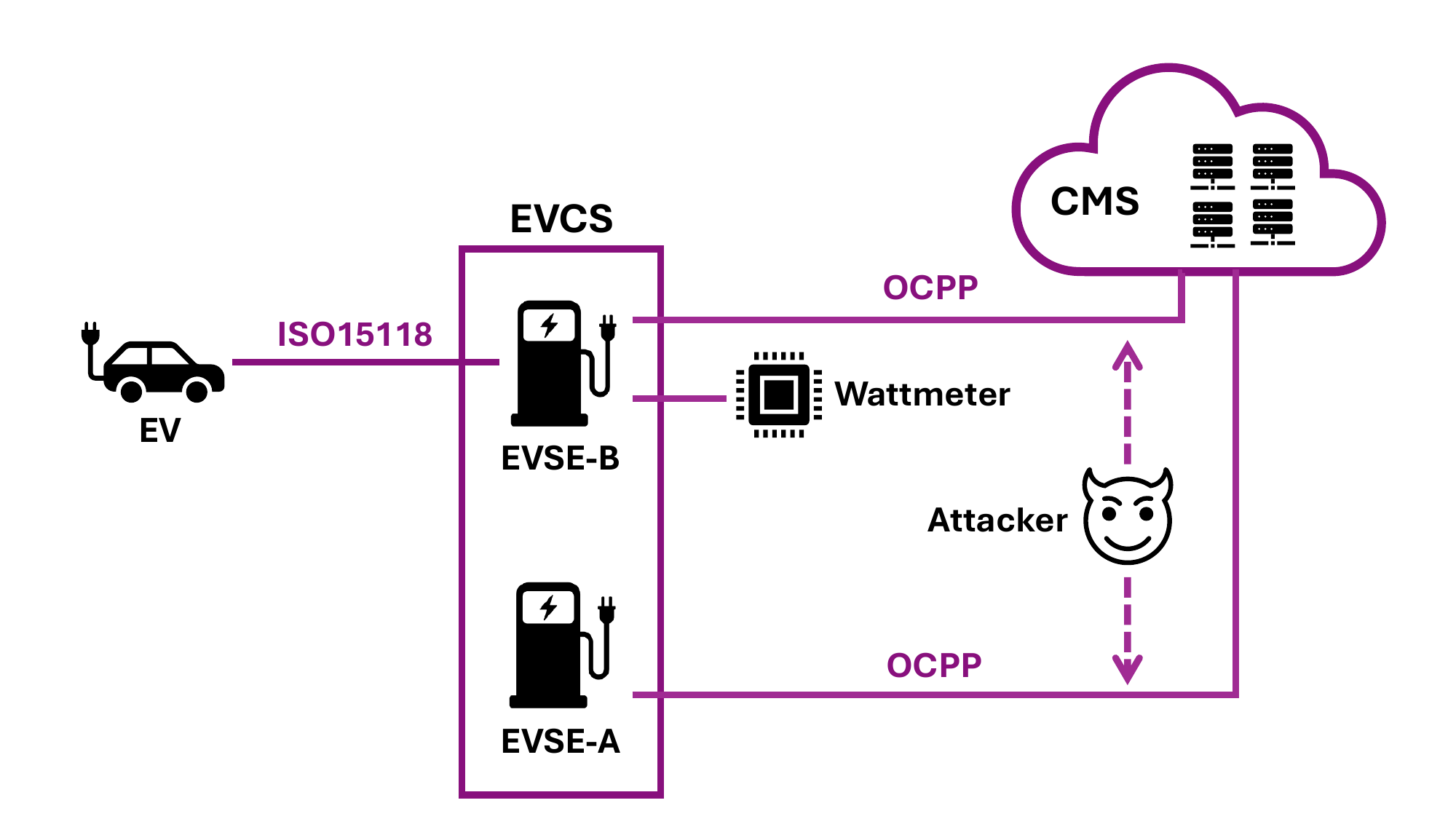}
\caption{Experimental setup of the EVSE dataset}
\label{fig:fig2}
\end{figure}

The network traffic dataset captures network traffic from two EVSE, referred to as EVSE-A and EVSE-B. EVSE-A communicates with the CMS via the Open Charge Point Protocol (OCPP), while EVSE-B connects to both an EV using ISO15118 protocol and the CSMS via OCPP. 
The dataset includes various types of Reconnaissance (Recon) and Denial of Service (DoS) cyberattacks, specifically targeting the OCPP interfaces of the supply equipment. Figure~\ref{fig:fig2} depicts the setup utilized for this dataset. 

\subsection{Preprocessings}

To extract network traffic data from the raw packet capture (pcap) files, the Python library NFStream is used. This process transformed the pcap data into CSV format, making it easier to analyze. Each of the initial datasets from district sources EVSE-A and EVSE-B contains 86 common features. 

Each charging station, EVSE-A and EVSE-B, has multiple CSV files, each corresponding to specific attack scenarios. To create a comprehensive dataset for each station, the individual CSV files are merged sequentially based on the time, using the initial timestamp of the bidirectional communication as the reference. This process ensures the proper chronological alignment of the data across all files. After merging the data, all timestamp-related features were removed to prevent temporal bias in the subsequent analysis. Since the datasets from EVSE-A and EVSE-B overlap in terms of time, they were kept separate to avoid interference, treating each station's data as a distinct entity.

Since the samples were not labeled initially, labels are added to the dataset based on the type of attack each sample was experiencing. Attack labels are assigned using the MAC addresses of the samples, as the attacker had a known MAC address in the simulated environment. Additionally, the type of attack is inferred from the CSV file names and included in the dataset. Following this, all features related to MAC and IP addresses, whether for the source or destination, are removed to avoid unnecessary bias, as these attributes may introduce location or network-specific information that could skew model predictions~\cite{10.1145/3457607}. It is anticipated that ML algorithms will classify and recognize data traffic based on statistical features such as inter-arrival times, packet length distribution, and the mean number of flows, which are more relevant for generalizing across different environments~\cite{DIMAURO2021104216}.

To categorize the output labels, a mapping is defined between the existing detailed labels and broader categories: Benign, Recon, and DoS. Recon attacks involve scanning networks or systems to gather information for potential future attacks, while DoS attacks are characterized by system overloads, causing services to become unavailable~\cite{buedi2024enhancing}. The distribution of these labels for each EVSE is shown in Table~\ref{table:output_distribution_comparison}.

\begin{table}[h!]
\centering
\caption{Comparison of Multi-Class Label Distributions for Charging Stations A and B}
\label{table:output_distribution_comparison}
\begin{tabular}{|c|c|c|}
\hline
\textbf{Label} & \textbf{EVSE-A Percentage} & \textbf{EVSE-B Percentage} \\ \hline
DoS            & 84.50\%                      & 23.87\%                      \\ \hline
Recon          & 12.55\%                      & 49.66\%                      \\ \hline
Benign         & 2.95\%                       & 26.47\%                      \\ \hline
\end{tabular}
\end{table}

Features with more than 99\% missing values in the samples, as well as those containing a single constant value across all samples, were removed to exclude non-impactful features. 
The nominal features related to application names and categories are encoded using label encoding. Although label encoding carries the risk of introducing ordinal relationships—since each label is assigned a unique number, unlike dummy coding, which creates separate binary columns for each label—it helps to keep the dataset size manageable by avoiding the significant dimensionality increase and dataset sparsity caused by dummy coding~\cite{iglesias2015analysis}. To ensure that this potential ordinality does not negatively impact the model, explainability methods are applied in the final analysis. This step ensures that the contribution of these features to the output is minimal and does not introduce any bias.

To remove highly correlated features, the correlation matrix for features in each dataset (EVSE-A and EVSE-B) is first calculated. Pairs of features with a correlation above 0.95, indicating strong linear relationships, are identified. The occurrences of each feature involved in these pairs are counted and prioritized, keeping those that appear most frequently. For each pair, one feature is kept, while the other is removed to minimize redundancy. Finally, the common features identified from this process across both datasets are removed to ensure consistency in features between the two datasets. At the end, there are 49 features. No further encoding or scaling is done.

The processed datasets, labeled as EVSE-A and EVSE-B, along with the accompanying code, are publicly accessible on GitHub for further research.\footnote{\href{https://github.com/Western-OC2-Lab/EVCI-Pruning}{https://github.com/Western-OC2-Lab/EVCI-Pruning}}


\section{Methodology}
\subsection{Model Training and Hyperparameter Optimization}

By referring to the Table~\ref{table:output_distribution_comparison}, it becomes clear that the distribution of labels is not consistent between the two charging stations. While the ideal dataset would have a balanced percentage of benign labels, in this case, EVSE-B was chosen for training and EVSE-A for testing. If the process were reversed, training would be done on a significantly smaller number of benign samples, leading to poor model performance due to the unbalanced distribution. To address this issue, techniques such as oversampling, undersampling, or synthetic data generation can be employed to balance the dataset~\cite{joloudari2023effective}. This is also important to note that merging the datasets from both stations is not a viable option, as it would disrupt the time dependency inherent in the time series data.

Three different models are used in this study: an MLP as the baseline model~\cite{Goodfellow-et-al-2016}, an LSTM designed specifically for time series data~\cite{10.1162/neco.1997.9.8.1735}, and XGBoost, a powerful gradient boosting algorithm~\cite{10.1145/2939672.2939785}. 

To preserve the ordinality of the data while ensuring all labels are included, the validation set is not selected randomly. Instead, the last 20\% of each detailed attack group is chosen as the validation set. This approach maintains the sequence of events and prevents label-specific leakage into the training set, as future samples are only used for validation after training on past data for each specific label. 

HPO, which involves tuning model parameters within a specified budget, is conducted to achieve optimal or near-optimal performance~\cite{YANG2020295}. This process is essential for improving model accuracy and efficiency across various architectures~\cite{10570354}. For this purpose, Optuna is used to efficiently search for the best hyperparameter configurations. As a flexible optimization framework, Optuna employs a define-by-run approach, meaning the hyperparameter search space is constructed dynamically during runtime rather than being predefined~\cite{akiba2019optuna}.

\begin{table*}[h!]
\centering
\caption{Optimized hyperparameters and best values selected for each model.}
\begin{tabular}{|c|l|l|}
\hline
\textbf{Model}                  & \textbf{Optimized Hyperparameters}                                                                 & \textbf{Best Parameters Selected} \\ \hline
\textbf{Multi-Layer Perceptron (MLP)} & \begin{tabular}[c]{@{}l@{}} - Number of layers: [1, 8] (step=1)\\ - Neurons per layer: {[}16, 32, 64, 128{]} \end{tabular} & \begin{tabular}[c]{@{}l@{}} - Number of layers: 4 \\ - Neurons per layer: [16, 128, 64, 3] \end{tabular}\\ \hline
\textbf{Long Short-Term Memory (LSTM)} & \begin{tabular}[c]{@{}l@{}} - Number of layers: [1, 3] (step=1)\\ - Units per layer: {[}50, 75, 100{]} \end{tabular} & \begin{tabular}[c]{@{}l@{}} - Number of layers: 3 \\ - Units per layer: [100, 50, 3] \end{tabular} \\ \hline
\textbf{XGBoost}                & \begin{tabular}[c]{@{}l@{}} - Number of estimators: {[}100, 1000{]} (step=100) \\ - Max depth: {[}3, 10{]} (step=1) \\ - Learning rate: {[}0.01, 0.3{]} (log scale) \\ - Subsample: {[}0.5, 1.0{]} (step=0.1) \\ - Colsample\_bytree: {[}0.5, 1.0{]} (step=0.1) \\ - Gamma: {[}0, 5{]} (step=1) \\ - Min child weight: {[}1, 10{]} (step=1) \\ - Lambda: {[}0.0, 10.0{]} (step=0.1) \\ - Alpha: {[}0.0, 10.0{]} (step=0.1)\end{tabular} &  \begin{tabular}[c]{@{}l@{}} - Number of estimators: 500 \\ - Max depth: 5 \\ - Learning rate: 0.0736 \\ - Subsample: 0.6452 \\ - Colsample\_bytree: 0.6745 \\ - Gamma: 0.8604 \\ - Min child weight: 9 \\ - Lambda (reg\_lambda): 7.6702 \\ - Alpha (reg\_alpha): 8.0907 \end{tabular} \\ \hline
\end{tabular}
\label{tab:hyperparameters}
\end{table*}

Optuna is used for HPO across three different models: MLP, LSTM, and XGBoost. For each model, fixed hyperparameters such as the number of trials (set to 10) and evaluation metrics (accuracy, precision, recall, and F1 score) are used. Optuna's objective function is designed to maximize validation accuracy during the tuning process. Additionally, several other fixed hyperparameters are applied across the models, including a batch size of 32, 50 epochs, and the Adam optimizer for MLP and LSTM models, with a window size of 5 for LSTM and default settings for XGBoost. For MLP, ReLU is the activation function for hidden layers and softmax for output layer. Early stopping with a patience of 5 epochs monitors validation loss, and categorical cross-entropy is used as the loss function for neural networks. Table~\ref{tab:hyperparameters} presents the tuned hyperparameters and their best-selected values for each model, highlighting the most effective configurations identified during optimization.

\subsection{Feature Selection}

Resource constraints limit the data to features that contribute more significantly to detecting the output~\cite{iglesias2015analysis}. In the realm of network traffic anomaly detection, FS is a critical step to ensure both computational efficiency and high detection performance. This paper proposes using SHAP for FS in MLP and LSTM algorithms. SHAP understands feature importance by measuring the contribution of each feature to the model's output. Rooted in Shapley values from cooperative game theory, SHAP calculates the marginal contribution of each feature across all possible combinations, ensuring consistent and interpretable results. This model-agnostic approach works with any ML model, making it a popular tool for explaining predictions~\cite{lundberg2017unified, NOORCHENARBOO2025115177}.

The kernel-based SHAP method is used for our neural network models. For the MLP, 100 random samples from the training set are used as background data, with predictions made on 1,000 random samples from the test set. For the LSTM, 100 consecutive samples from the training set and 1,000 consecutive samples from the test set are selected to preserve the temporal sequence. In calculating SHAP values for the LSTM, an average is taken over the specified sample window size for each feature, allowing the SHAP values to represent the average contribution of each feature across multiple time steps. Increasing sample sizes could potentially improve accuracy, though this was not feasible due to resource constraints.

Finally, for XGBoost, inherent model explainability through feature importance is utilized to understand the influence of individual features on predictions. This built-in explainability provides valuable insights into which features have the greatest impact on the model’s outcomes.

\begin{figure}[ht]
\centering
\includegraphics[width=0.45\textwidth]{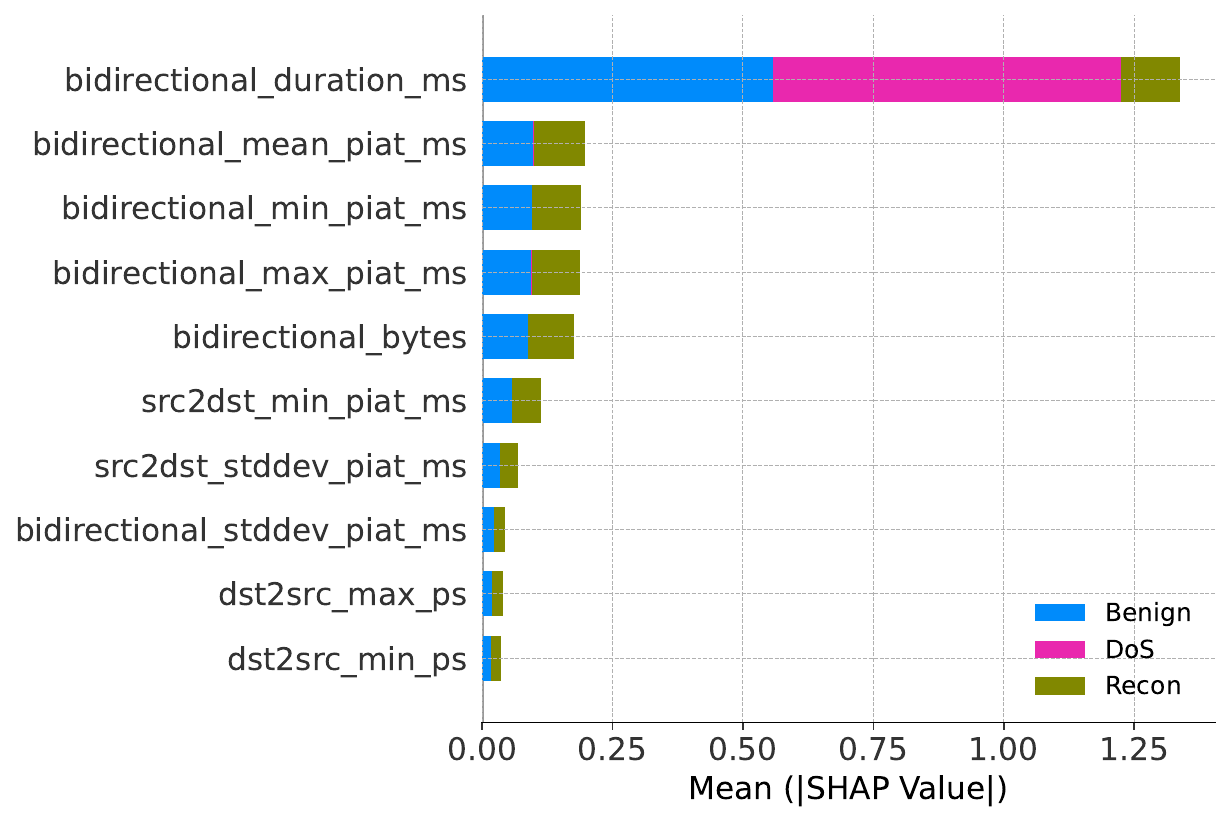}
\caption{SHAP Summary Plot for MLP Model}
\label{fig:mlp_shap}
\end{figure}
\begin{figure}[ht]
\centering
\includegraphics[width=0.45\textwidth]{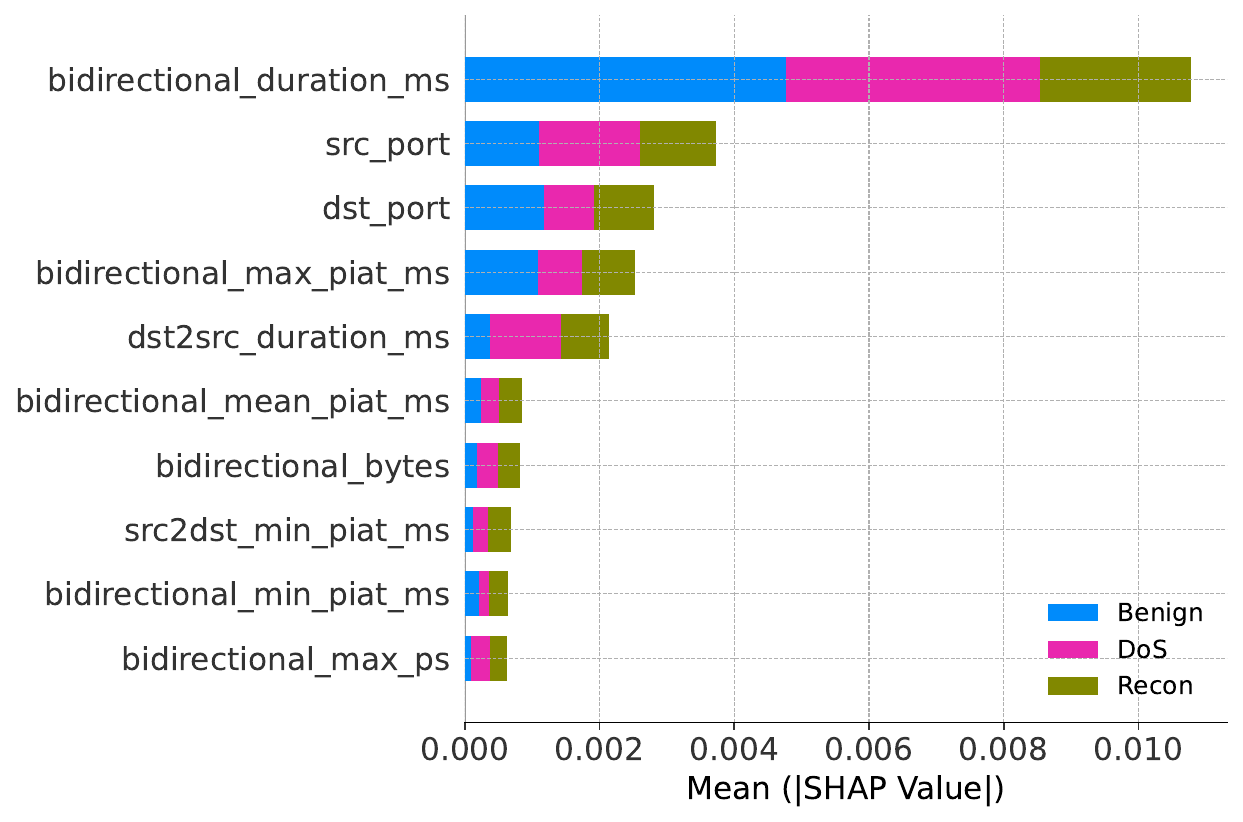}
\caption{SHAP Summary Plot for LSTM Model}
\label{fig:lstm_shap}
\end{figure}
\begin{figure}[ht]
\centering
\includegraphics[width=0.45\textwidth]{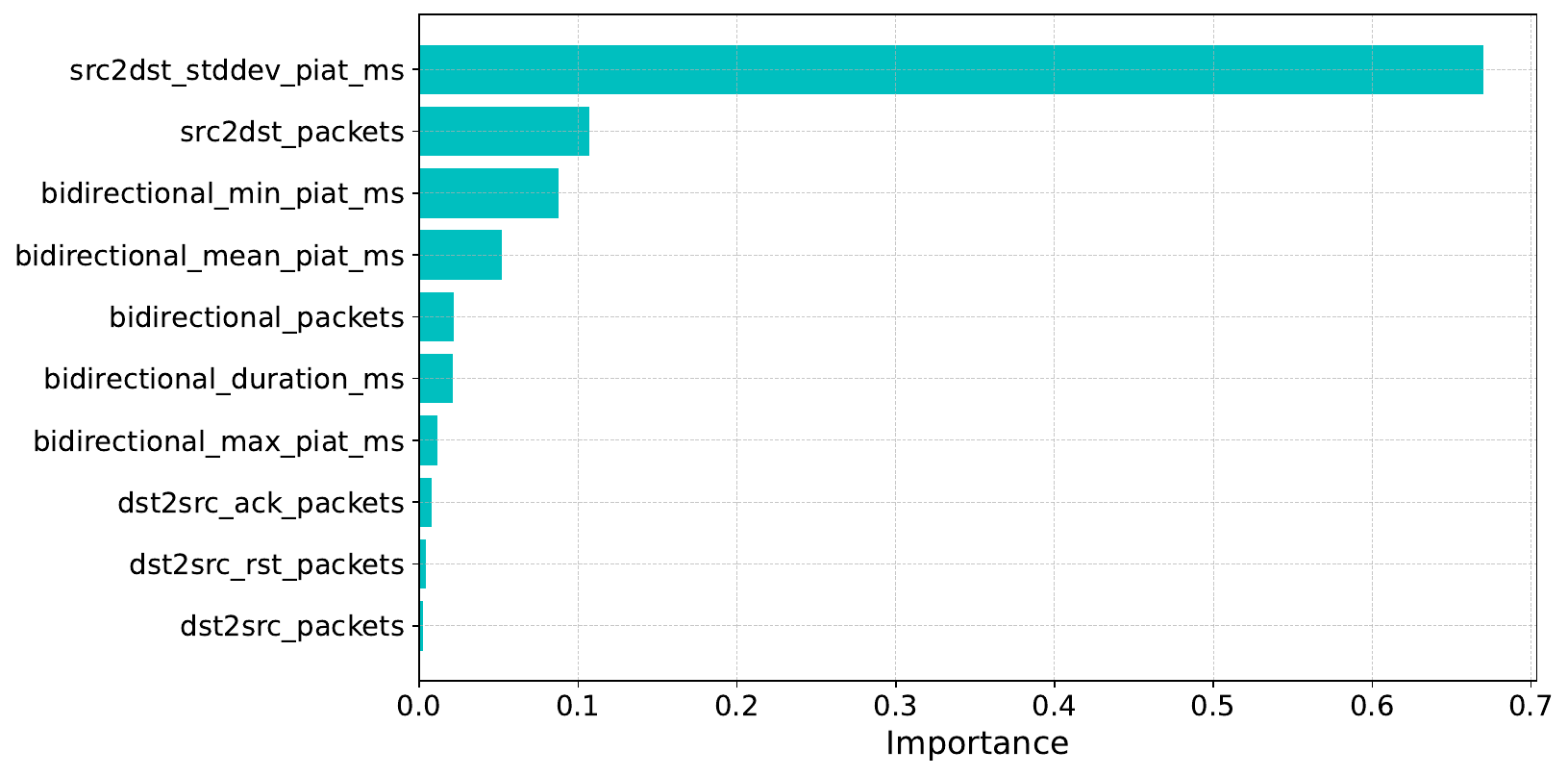}
\caption{Feature Importance Plot for XGBoost Model}
\label{fig:xgb_plot}
\end{figure}

Figures~\ref{fig:mlp_shap} and~\ref{fig:lstm_shap} display the average absolute SHAP values for the top ten features, representing the average impact of each feature on the model's output magnitude. Each SHAP value indicates the average change in the model’s prediction due to that feature, measured in the model's output units. The higher the absolute SHAP value, the greater the feature's influence on the prediction outcome, regardless of direction. Furthermore, figure~\ref{fig:xgb_plot} illustrates the ten top important features derived from the XGBoost model, highlighting their individual contributions to the model's predictive accuracy.

\subsection{Model Pruning}

\begin{table*}[h]
\centering
\caption{Evaluation Metrics for MLP, LSTM, and XGBoost Models at Different Stages}
\label{table:combined_results}
\begin{tabular}{|c|c|c|c|c|c|c|c|}
\hline
\textbf{Model Name} & \textbf{Model Stage} & \textbf{Accuracy} & \textbf{Precision} & \textbf{Recall} & \textbf{F1 Score} & \textbf{Average Inference Time (ms)} & \textbf{Model Size (KB)} \\ \hline

\multirow{3}{*}{MLP} & Original& 0.9841 & 0.9858 & 0.9841 & 0.9846 & 63.704041 & 46.70 \\ 
 & Pruned & 0.9804 & 0.9854 & 0.9804 & 0.9817 & 3.016179 & 36.46 \\ 
 & Feature-Selected Pruned & 0.9835 & 0.9868 & 0.9835 & 0.9843 & 1.317868 & 34.60 \\ \hline

\multirow{3}{*}{LSTM} & Original & 0.9915 & 0.9919 & 0.9915 & 0.9916 & 63.612781 & 355.07 \\ 
 & Pruned & 0.9919 & 0.9923 & 0.9919 & 0.9920 & 24.735132 & 253.64 \\ 
 & Feature-Selected Pruned & 0.9921 & 0.9925 & 0.9921 & 0.9922 & 9.948283 & 210.84 \\ \hline

\multirow{3}{*}{XGBoost} & Original & 0.9850 & 0.9865 & 0.9850 & 0.9854 & 0.202661 & 670.566 \\ 
 & Pruned & 0.9815 & 0.9845 & 0.9815 & 0.9823 & 0.198487 & 219.632 \\ 
 & Feature-Selected Pruned& 0.9815 & 0.9844 & 0.9815 & 0.9823 & 0.196298 & 216.716 \\ \hline

\end{tabular}
\end{table*}

For model pruning in neural network-based models, specifically MLP and LSTM in this paper, a gradual pruning technique introduced by Zhu and Gupta~\cite{zhu2017prune} is employed to incrementally set low-magnitude weights to zero. This method increases the sparsity level over time, governed by a cubic decay function in the sparsity update formula. This approach balances model compression with performance by allowing the network to adapt to each pruning step, minimizing the impact on accuracy.

The final sparsity target for these neural network-based models is set at 65\%, meaning that 65\% of low-magnitude weights are pruned to zero while the remaining weights are fine-tuned to preserve accuracy. A key challenge at this stage is that pruning alone does not reduce model size or inference time, as each zero in the weight matrix is still stored as a 32-bit float and processed like any other value. Consequently, predictions continue to perform operations on these zeroed weights, offering no improvement in computational efficiency.

To address the inefficiency of storing and processing zero values, all operations are performed in a sparse format. In this format, matrices are converted to store only non-zero values, dramatically reducing memory usage. Specifically, a compressed sparse row (CSR) format is used, where each matrix is represented by three arrays: one for the non-zero values, one indicating the column index of each value, and a third row-pointer array that marks the start of each row in the values array. This structure allows efficient storage and computation by eliminating the need to store or process zeros~\cite{parashar2017scnn}.

To achieve efficient prediction in neural networks using sparse matrices, custom inference functions are developed to maintain sparsity throughout computations. For MLPs, this approach involves converting dense layer weights to sparse matrices, performing matrix multiplications with the input, and adding biases. Sparse activations, such as ReLU and softmax, are applied exclusively to non-zero elements, thus preserving sparsity across layers. In contrast, inference for LSTMs presents a greater challenge due to their sequential architecture. The LSTM inference function constructs individual sparse matrices for each gating mechanism---input, forget, cell, and output---enabling precise matrix multiplications and updates to the cell and hidden states at each timestep.

Pruning of the MLP and LSTM models can be further enhanced through FS. In these models, FS involves removing input nodes along with their connected layers, thereby contributing to further weight reduction and pruning. Based on SHAP values, the top 10 contributing features were identified as the most influential candidates. In the MLP model, these features account for approximately 90\% of the total feature importance, while in the LSTM model, they contribute around 80\% to the overall feature importance.

For XGBoost, the number of estimators and maximum depth were reduced from 500 to 100 and from 5 to 2, respectively, to create a more compact model. By selecting the top 10 features based on XGBoost's inherent feature importance, over 98\% of the model's total feature importance is retained.

The chosen values for sparsity target, number of estimators, maximum depth, and number of features represent a balance between compressing the model and retaining acceptable levels of accuracy. This approach provides significant improvements in computational efficiency and memory usage, with the primary aim of minimizing any noticeable drop in evaluation metrics.

\section{Results and Analysis}

Each model is investigated in three stages: first, the original model itself; second, the model with a pruned structure; and third, the model with both FS and pruning to further reduce its size. These pruned models are compared with the original versions in terms of evaluation metrics, model size, and inference time. All experiments were conducted on a high-performance desktop with a 12th Gen Intel Core i9 processor and 64 GB of RAM, running Microsoft Windows 10 to ensure consistency in evaluating model size and inference time across all stages. Table~\ref{table:combined_results} presents the results for the MLP, LSTM, and XGBoost models.

The results for both the MLP and LSTM models show improvements in model efficiency through pruning and FS, with minimal impact on key evaluation metrics. In terms of accuracy, precision, recall, and F1 score, both models maintain high performance across the original, pruned, and feature-selected pruning stages. Accuracy is calculated over all samples, while precision, recall, and F1 score are averaged across classes using a weighted approach, reflecting each class's contribution based on its frequency in the dataset.

A major impact of pruning and FS is observed in inference time and model size. For these neural network models, 10\% of the test set is randomly selected, and the average inference time per sample is measured.
This subset, containing more than 54,000 samples, is sufficiently large to reflect the behavior of the entire dataset while reducing the computational burden caused by the test set's size. 
For the MLP model, the average inference time per sample decreases substantially from 63.70 ms in the original model to 1.32 ms in the feature-selected pruned model, a reduction of around 98\%. The LSTM model also benefits significantly from these optimizations, with inference time dropping from 63.61 ms to 9.95 ms, a reduction of approximately 84.4\%, demonstrating how sparsity effectively reduces computational load.

Model sizes are presented here based solely on the weight parameters, as these are directly impacted by pruning and feature optimization. The original model's weights are saved in a dense format, while the pruned models' weights are stored in sparse format to reflect the reduced parameters more accurately. For the MLP, the model size decreases from 46.70 KB to 34.60 KB, while the LSTM sees a considerable reduction from 355.07 KB to 210.84 KB. These reductions indicate that pruning and FS not only improve inference speed but also optimize memory usage, making both models more efficient and scalable without significant loss in accuracy.

For the XGBoost model, pruning and FS achieve a significant reduction in model size, along with a decrease in inference time, with minimal impact on performance. The average inference time across all test set samples decreases from 0.203 ms in the original model to 0.196 ms in the pruned model with FS, improving computational efficiency. Model size is also reduced from 670.57 KB to 216.72 KB, highlighting the impact of these techniques on memory usage.

Through the pruning and FS techniques applied in this work, model size and inference time are significantly reduced, making it feasible to implement these models on resource-constrained, low-power hardware typical of TinyML applications. This enables TinyML for real-time anomaly detection in EVs, where efficient model deployment is crucial for maintaining performance while minimizing computational and memory demands.

\section{Conclusion}
This study demonstrates the effectiveness of pruning and FS for deploying efficient ML models in EVCI. Using the CICEVSE2024 dataset, we optimize three models—MLP, LSTM, and XGBoost—through unstructured pruning and SHAP-based FS. As a result, inference time is reduced by 97.93\% for MLP, 84.36\% for LSTM, and 3.14\% for XGBoost, while model sizes decrease by 25.91\%, 40.62\%, and 67.68\%, respectively, with performance metrics dropping by less than 0.5\% across all models. These results highlight the potential of targeted model compression techniques for deploying high-performance, low-latency models in environments with limited memory and processing resources. This research paves the way for future work in TinyML and model compression techniques, emphasizing the feasibility of on-device, real-time anomaly detection in EVCI and similar IoT domains.

\bibliographystyle{ieeetr}
\bibliography{References}

\begin{thebibliography}{10}

\bibitem{menghani2023efficient}
G.~Menghani, ``Efficient deep learning: A survey on making deep learning models smaller, faster, and better,'' {\em ACM Computing Surveys}, vol.~55, no.~12, pp.~1--37, 2023.

\bibitem{patil2023federated}
D.~Patil, S.~Goel, S.~K. Garud, M.~D. Kokate, A.~Nashte, and P.~Rane, ``Federated learning in real-time medical iot: Optimizing privacy and accuracy for chronic disease monitoring.,'' {\em Journal of Electrical Systems}, vol.~19, no.~3, 2023.

\bibitem{gamanayake2020cluster}
C.~Gamanayake, L.~Jayasinghe, B.~K.~K. Ng, and C.~Yuen, ``Cluster pruning: An efficient filter pruning method for edge ai vision applications,'' {\em IEEE Journal of Selected Topics in Signal Processing}, vol.~14, no.~4, pp.~802--816, 2020.

\bibitem{samal2020attention}
K.~Samal, M.~Wolf, and S.~Mukhopadhyay, ``Attention-based activation pruning to reduce data movement in real-time ai: A case-study on local motion planning in autonomous vehicles,'' {\em IEEE Journal on Emerging and Selected Topics in Circuits and Systems}, vol.~10, no.~3, pp.~306--319, 2020.

\bibitem{10.1145/3657282}
H.-I. Liu, M.~Galindo, H.~Xie, L.-K. Wong, H.-H. Shuai, Y.-H. Li, and W.-H. Cheng, ``Lightweight deep learning for resource-constrained environments: A survey,'' {\em ACM Comput. Surv.}, vol.~56, jun 2024.

\bibitem{lecun1989optimal}
Y.~LeCun, J.~Denker, and S.~Solla, ``Optimal brain damage,'' {\em Advances in neural information processing systems}, vol.~2, 1989.

\bibitem{abadade2023comprehensive}
Y.~Abadade, A.~Temouden, H.~Bamoumen, N.~Benamar, Y.~Chtouki, and A.~S. Hafid, ``A comprehensive survey on tinyml,'' {\em IEEE Access}, 2023.

\bibitem{10620263}
F.~Dehrouyeh, L.~Yang, F.~Badrkhani~Ajaei, and A.~Shami, ``On tinyml and cybersecurity: Electric vehicle charging infrastructure use case,'' {\em IEEE Access}, vol.~12, pp.~108703--108730, 2024.

\bibitem{cheng2018model}
Y.~Cheng, D.~Wang, P.~Zhou, and T.~Zhang, ``Model compression and acceleration for deep neural networks: The principles, progress, and challenges,'' {\em IEEE Signal Processing Magazine}, vol.~35, no.~1, pp.~126--136, 2018.

\bibitem{ullrich2017soft}
K.~Ullrich, E.~Meeds, and M.~Welling, ``Soft weight-sharing for neural network compression,'' {\em arXiv preprint arXiv:1702.04008}, 2017.

\bibitem{unbdataset}
{University of New Brunswick}, ``Evse dataset 2024.'' \url{https://www.unb.ca/cic/datasets/evse-dataset-2024.html}.
\newblock Accessed: April 29, 2024.

\bibitem{buedi2024enhancing}
E.~D. Buedi, A.~A. Ghorbani, S.~Dadkhah, and R.~L. Ferreira, ``Enhancing ev charging station security using a multi-dimensional dataset: Cicevse2024,'' 2024.

\bibitem{10.1145/3457607}
N.~Mehrabi, F.~Morstatter, N.~Saxena, K.~Lerman, and A.~Galstyan, ``A survey on bias and fairness in machine learning,'' {\em ACM Comput. Surv.}, vol.~54, jul 2021.

\bibitem{DIMAURO2021104216}
M.~{Di Mauro}, G.~Galatro, G.~Fortino, and A.~Liotta, ``Supervised feature selection techniques in network intrusion detection: A critical review,'' {\em Engineering Applications of Artificial Intelligence}, vol.~101, p.~104216, 2021.

\bibitem{iglesias2015analysis}
F.~Iglesias and T.~Zseby, ``Analysis of network traffic features for anomaly detection,'' {\em Machine Learning}, vol.~101, pp.~59--84, 2015.

\bibitem{joloudari2023effective}
J.~H. Joloudari, A.~Marefat, M.~A. Nematollahi, S.~S. Oyelere, and S.~Hussain, ``Effective class-imbalance learning based on smote and convolutional neural networks,'' {\em Applied Sciences}, vol.~13, no.~6, p.~4006, 2023.

\bibitem{Goodfellow-et-al-2016}
I.~Goodfellow, Y.~Bengio, and A.~Courville, {\em Deep Learning}.
\newblock MIT Press, 2016.
\newblock \url{http://www.deeplearningbook.org}.

\bibitem{10.1162/neco.1997.9.8.1735}
S.~Hochreiter and J.~Schmidhuber, ``{Long Short-Term Memory},'' {\em Neural Computation}, vol.~9, pp.~1735--1780, 11 1997.

\bibitem{10.1145/2939672.2939785}
T.~Chen and C.~Guestrin, ``Xgboost: A scalable tree boosting system,'' in {\em Proceedings of the 22nd ACM SIGKDD International Conference on Knowledge Discovery and Data Mining}, KDD '16, (New York, NY, USA), p.~785–794, Association for Computing Machinery, 2016.

\bibitem{YANG2020295}
L.~Yang and A.~Shami, ``On hyperparameter optimization of machine learning algorithms: Theory and practice,'' {\em Neurocomputing}, vol.~415, pp.~295--316, 2020.

\bibitem{10570354}
I.~Shaer, S.~Nikan, and A.~Shami, ``Efficient transformer-based hyper-parameter optimization for resource-constrained iot environments,'' {\em IEEE Internet of Things Magazine}, vol.~7, no.~6, pp.~102--108, 2024.

\bibitem{akiba2019optuna}
T.~Akiba, S.~Sano, T.~Yanase, T.~Ohta, and M.~Koyama, ``Optuna: A next-generation hyperparameter optimization framework,'' in {\em Proceedings of the 25th ACM SIGKDD international conference on knowledge discovery \& data mining}, pp.~2623--2631, 2019.

\bibitem{lundberg2017unified}
S.~Lundberg, ``A unified approach to interpreting model predictions,'' {\em arXiv preprint arXiv:1705.07874}, 2017.

\bibitem{NOORCHENARBOO2025115177}
M.~Noorchenarboo and K.~Grolinger, ``Explaining deep learning-based anomaly detection in energy consumption data by focusing on contextually relevant data,'' {\em Energy and Buildings}, vol.~328, p.~115177, 2025.

\bibitem{zhu2017prune}
M.~Zhu and S.~Gupta, ``To prune, or not to prune: exploring the efficacy of pruning for model compression,'' {\em arXiv preprint arXiv:1710.01878}, 2017.

\bibitem{parashar2017scnn}
A.~Parashar, M.~Rhu, A.~Mukkara, A.~Puglielli, R.~Venkatesan, B.~Khailany, J.~Emer, S.~W. Keckler, and W.~J. Dally, ``Scnn: An accelerator for compressed-sparse convolutional neural networks,'' {\em ACM SIGARCH computer architecture news}, vol.~45, no.~2, pp.~27--40, 2017.

\end{thebibliography}

\end{document}